# LoRA-fine-tuned Large Vision Models for Automated Assessment of Post-SBRT Lung Injury

M. Bolhassani[1], B. Veasey[1], E. Daugherty[2], S. Keltner[2],

N. Kumar[2], N. Dunlap[3], A. Amini[1], Fellow, IEEE

[1]Electrical and Computer Engineering Department, University of Louisville, KY
[2]Department of Radiation Oncology, University of Cincinnati College of Medicine, Cincinnati, OH
[3]Department of Radiation Oncology, University of Louisville Hospital, Louisville, KY

*Abstract*— This study investigates the efficacy of Low-Rank Adaptation (LoRA) for fine-tuning large Vision Models, DinoV2 and SwinV2, to diagnose Radiation-Induced Lung Injury (RILI) from X-ray CT scans following Stereotactic Body Radiation Therapy (SBRT). To evaluate the robustness and efficiency of this approach, we compare LoRA with traditional full fine-tuning and inference-only (no fine-tuning) methods. Cropped images of two sizes (50 $mm^3$ and 75 $mm^3$), centered at the treatment isocenter, in addition to different adaptation techniques for adapting the 2D LVMs for 3D data were used to determine the sensitivity of the models to spatial context. Experimental results show that LoRA achieves comparable or superior performance to traditional fine-tuning while significantly reducing computational costs and training times by requiring fewer trainable parameters.

*Clinical Relevance*— This study improves the detection of Radiation-Induced Lung Injury (RILI) in lung cancer patients following SBRT, enabling AI-driven diagnosis to support clinical decision making.

## I. Introduction

Radiation-Induced Lung Injury (RILI) is a significant dose-limiting complication in lung cancer patients undergoing Stereotactic Body Radiation Therapy (SBRT), with reported incidences ranging from 5 to 25% [1]. Diagnosing RILI is challenging due to overlapping symptoms with other pulmonary conditions and the evolving nature of radiographic features. These features progress from subtle ground-glass attenuation in early stages to more pronounced linear scarring in advanced phases [2]. While Convolutional Neural Networks (CNNs) have shown promise in classifying RILI with acceptable accuracy [3], recent advances in Vision Transformers (ViTs) [4] offer an opportunity to explore whether these models can further improve diagnostic performance. Transformers, with their ability to capture global context and long-range dependencies, have already demonstrated exceptional performance in natural image processing tasks. Adapting pretrained ViTs to medical imaging, particularly for RILI classification, presents an opportunity but also challenges including domain differences and the need for reducing computational costs.

In this study, we investigate the performance of state-of-the-art Vision Transformers, DinoV2 (Self-Distillation with NO labels) ViT [5] and SwinV2 transformers [6], fine-tuned with Low-Rank Adaptation (LoRA) [7] to classify RILI from post-SBRT X-ray CT scans. We use cropped images of two sizes (50 $mm^3$ and 75 $mm^3$) centered at the treatment isocenter, to compare LoRA-based fine-tuning with traditional fine-tuning and inference-only approaches, assessing both diagnostic accuracy and computational efficiency.

This study extends our previous work [3], which explored post-SBRT radiation-induced lung injury diagnosis using Convolutional Neural Networks, and [8], who applied radiomics to distinguish RILI from recurrent cancer in post-SBRT follow-up CTs using a more limited dataset of 22 subjects. Other studies [8-11] have utilized radiomics for predicting RILI from pre-treatment scans. However, this study is the first to explore the adaptation of large vision models for RILI classification in post-SBRT scans in a larger cohort.

## II. Data

The dataset for this study consisted of 138 follow-up CT scans from 41 unique subjects across two institutions, focusing on treated nodules with an average size of 2.6 cm. For validation, an independent holdout test set comprising 83 follow-up scans from 26 subjects. This test set represents a range of treatment dosages and includes scans collected using different CT scanners [3]. Figure 1 shows exemplar positive and negative samples of RILI dataset, with red boxes highlighting the affected regions of interest (ROIs).

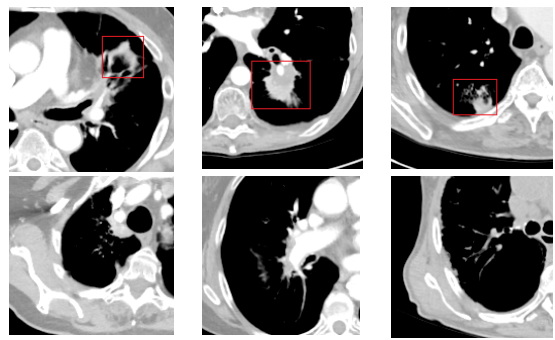

Figure 1. Examples of positive (top row) and negative (bottom row) Radiation-Induced Lung Injury (RILI) sample CT images are shown. The rectangles in the top row are centered around the treatment isocenter and enclose the areas affected by RILI.

To prepare the data for deep learning model training and testing, several preprocessing steps were applied. First, all CT scans were resampled to a standardized resolution of [1, 1, 2] mm/voxel to account for variations in image resolution across scanners and institutions. Each follow-up CT scan was then aligned with its corresponding pre-treatment scan. After alignment, an isometric volume of either 50 mm³ or 75 mm³, centered at the treatment isocenter, was cropped [3]. The resulting ROIs were extracted and windowed to a range of -500 to 200 HU. These ROIs were then normalized to a scale of 0 to 1 using linear scaling, which enhances the visibility of Hounsfield Unit (HU) values typical of lung tissue and improves model training efficiency [3].

The dataset was partitioned at the patient level to avoid data leakage, ensuring that no patient appeared in both training and test sets. The data was initially split into 60% for training and 40% for testing. The training set was then further divided, with 80% used for training and 20% reserved for validation.

## III. METHOD

This study analyzed data from lung cancer patients who underwent SBRT at two specialized cancer centers. The focus of the study was on applying domain adaptation techniques with pre-trained large vision models, such as DINOV2 and SwinV2 Transformers, to improve early diagnostic capabilities for RILI. The dataset, which included 67 subjects, was divided into training and holdout test sets, with careful management to prevent data leakage between training and testing phases. Additionally, to evaluate the influence of contextual information on model performance, separate training iterations were performed using 50 mm³ and 75 mm³ volumes centered on the treated region.

To evaluate the domain adaptation of pre-trained Vision Transformers (ViTs), we tested 48 models across combinations of 2 input configurations, 2 crop sizes, 3 adaptation techniques, and 4 model architectures. This setup varied key parameters, such as input configuration and fine-tuning methods, while keeping the datasets and validation methods consistent. Further details are provided in subsequent sections. An overview of large vision models explored in this paper is presented in Table I.

TABLE I. LARGE VISION MODELS STUDIED

| Name | Network Architecture | No. of Params |
|---|---|---|
| DINOv2-s | ViT | 22M |
| DINOv2-b | ViT | 86.5M |
| SwinV2-s | Swin | 49M |
| SwinV2-b | Swin | 87.9M |

Abbreviations: "-s" denotes small models, while "-b" refers to base models of the pre-trained architectures

### A. Input configurations

To effectively adapt 2D LVMs networks to 3D data, the input was structured to capture the SBRT-treated region. Specifically, three orthogonal slices—mid-axial, mid-sagittal, and mid-coronal—were extracted from the centroid of the treated area in the RILI dataset. These slices were then combined into three input channels, preserving key 3D spatial information in a 2D format. For comparison, a single mid-axial slice repeated three times for the three channels was also utilized to evaluate the impact of incorporating 3D context.

### B. Domain Adaptation Methods

To adapt DINOv2 and SwinV2 Vision Transformers, originally pre-trained on natural images, to the RILI domain, fine-tuning on the new, limited dataset is required. LoRA fine-tuning incorporates a low-rank adapter matrix into each attention layer of transformer-based models, such as ViT-DINOv2 and ViT-SwinV2. As shown in Figure 2, the LoRA module modifies the original model weights by introducing these low-rank matrices. We compare the following three fine-tuning techniques:

- No Fine-Tuning (NFT): This baseline approach keeps the ViT backbone unchanged, updating only the weights of the linear classification layer.

- Full Fine-Tuning (FFT): The standard approach for fine-tuning transformers, which involves fine-tuning of all layers of the model.

- Low Rank Adaptation (LoRA): Parameter-efficient fine-tuning method evaluated in this study, designed to reduce the number of trainable parameters, accelerate training, and improve classification performance (see Fig. 2).

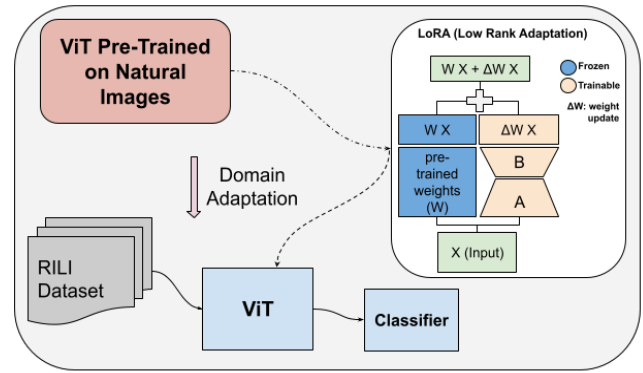

Figure 2. Domain adaptation for RILI diagnosis using LoRA [7]. LoRA efficiently fine-tunes large Vision Transformers (ViTs) by updating only a small subset of task-specific parameters while freezing pre-trained weights. This preserves the rich representations learned from natural images.

Additionally, Table II provides a comprehensive comparison of training time per epoch, and trainable parameter counts across the three tuning methods for the top-performing model (DINOv2-base with orthogonal input), revealing that LoRA substantially reduces both parameter count and training time relative to the fully fine-tuning method.

TABLE II. IMPELEMENTATION DETAILS FOR DINOV2 BASE MODEL

| Model | Tuning | Trainable Params | Training Time/epoch (sec) |
|---|---|---|---|
| DINO V2- base (3D Orthogonal Inputs) | NF | 3074 | 2.35 |
| | FFT | 86.5 M | 5.17 |
| | LoRA | 1.2 M | 3.07 |

TABLE III. PERFORMANCE COMPARISON OF DINO V2 BASE MODELS WITH DIFFERENT TUNING METHODS

| Model | | Crop Size mm³ | ROC-AUC | F1 | Precision | Recall | Specificity | Accuracy |
|---|---|---|---|---|---|---|---|---|
| 3D ResNet-10[3] | | 50 | 0.762 ± 0.022 | 0.789 ± 0.043 | 0.868 ± 0.023 | 0.729 ± 0.082 | 0.663 ± 0.098 | |
| | | 75 | 0.693 ± 0.038 | 0.726 ± 0.039 | 0.816 ± 0.020 | 0.657 ± 0.064 | 0.559 ± 0.086 | |
| DINO V2-base (2D Axial Inputs) | NFT | 50 | 0.522±.11 | 0.445±.2 | 0.756±.15 | 0.377±.31 | 0.62±.3 | 0.43±.17 |
| | | 75 | 0.633±.07 | 0.534±.22 | 0.883±.1 | 0.452±.3 | 0.7±.36 | 0.513±.14 |
| | FFT | 50 | 0.556±.058 | 0.53±.34 | 0.821±.1 | 0.518±.41 | 0.54±.3 | 0.532±.21 |
| | | 75 | 0.625±.08 | 0.518±.31 | 0.818±.1 | 0.504±.4 | 0.56±.37 | 0.518±.22 |
| | LoRA | 50 | 0.626±.12 | 0.53±.29 | 0.816±.04 | 0.455±.3 | 0.71±.2 | 0.518±.18 |
| | | 75 | **0.728±.04** | 0.587±.33 | 0.872±.08 | 0.54±.3 | 0.69±.27 | 0.577±.22 |
| DINO V2-base (3D Orthogonal Inputs) | NFT | 50 | 0.645±.052 | 0.542±.09 | 0.859±.055 | 0.458±.11 | 0.714±.12 | 0.522±.077 |
| | | 75 | 0.715±.05 | 0.623±.07 | 0.887±.042 | 0.567±.09 | 0.715±.1 | 0.604±.06 |
| | FFT | 50 | 0.696±.065 | 0.75±.09 | 0.828±.04 | 0.704±.15 | 0.53±.18 | 0.661±.072 |
| | | 75 | **0.79±.081** | **0.861±.0.02** | 0.847±.02 | 0.875±.035 | 0.52±.067 | **0.787±.034** |
| | LoRA | 50 | 0.736±.057 | 0.718±.141 | 0.838±.037 | 0.659±.22 | 0.6±.19 | 0.64±.125 |
| | | 75 | **0.783±.069** | 0.764±.13 | 0.852±.059 | 0.724±.216 | 0.59±.0.23 | **0.711±.126** |

TABLE IV. PERFORMANCE COMPARISON OF DINO V2 SMALL MODELS WITH DIFFERENT TUNING METHODS

| Model | | Crop Size mm³ | ROC-AUC | F1 | Precision | Recall | Specificity | Accuracy |
|---|---|---|---|---|---|---|---|---|
| 3D ResNet-10[3] | | 50 | **0.762 ± 0.022** | **0.789 ± 0.043** | 0.868 ± 0.023 | 0.729 ± 0.082 | 0.663 ± 0.098 | |
| | | 75 | 0.693 ± 0.038 | 0.726 ± 0.039 | 0.816 ± 0.020 | 0.657 ± 0.064 | 0.559 ± 0.086 | |
| DINO V2-small (2D Axial Inputs) | NFT | 50 | 0.608±.04 | 0.562±.27 | 0.837±.09 | 0.491±.29 | 0.64±.22 | 0.528±.17 |
| | | 75 | 0.636±.09 | 0.59±.29 | 0.858±.08 | 0.524±.29 | 0.66±.22 | 0.558±.17 |
| | FFT | 50 | 0.608±.047 | 0.703±.25 | 0.789±.019 | 0.714±.3 | 0.4±.28 | 0.637±.17 |
| | | 75 | 0.628±.06 | 0.718±.22 | 0.774±.017 | 0.734±.3 | 0.36±.26 | 0.641±.26 |
| | LoRA | 50 | 0.642±.06 | 0.581±.26 | **0.884±.1** | 0.518±.3 | 0.67±.3 | 0.555±.16 |
| | | 75 | 0.692±.04 | 0.691±.16 | 0.852±.069 | 0.632±.25 | 0.6±.29 | 0.624±.12 |
| DINO V2-small (3D Orthogonal Inputs) | NFT | 50 | 0.677±.09 | 0.656±.08 | 0.826±.04 | 0.55±.1 | 0.65±.1 | 0.575±.08 |
| | | 75 | **0.725±.04** | 0.687±.06 | 0.856±.01 | 0.58±.1 | 0.7±.079 | 0.609±.057 |
| | FFT | 50 | 0.711±.02 | 0.747±.07 | 0.831±.02 | 0.688±.12 | 0.57±.12 | 0.659±.07 |
| | | 75 | 0.703±.02 | 0.791±.07 | 0.849±.01 | 0.75±.12 | 0.59±.1 | **0.711±.07** |
| | LoRA | 50 | 0.671±.03 | 0.784±.04 | 0.845±.02 | 0.737±.09 | 0.58±.13 | 0.698±.04 |
| | | 75 | **0.721±.06** | **0.789±.08** | 0.839±.02 | 0.754±.13 | 0.56±.1 | **0.706±.08** |

*Full version is under EMBC 2025 Conference publication*

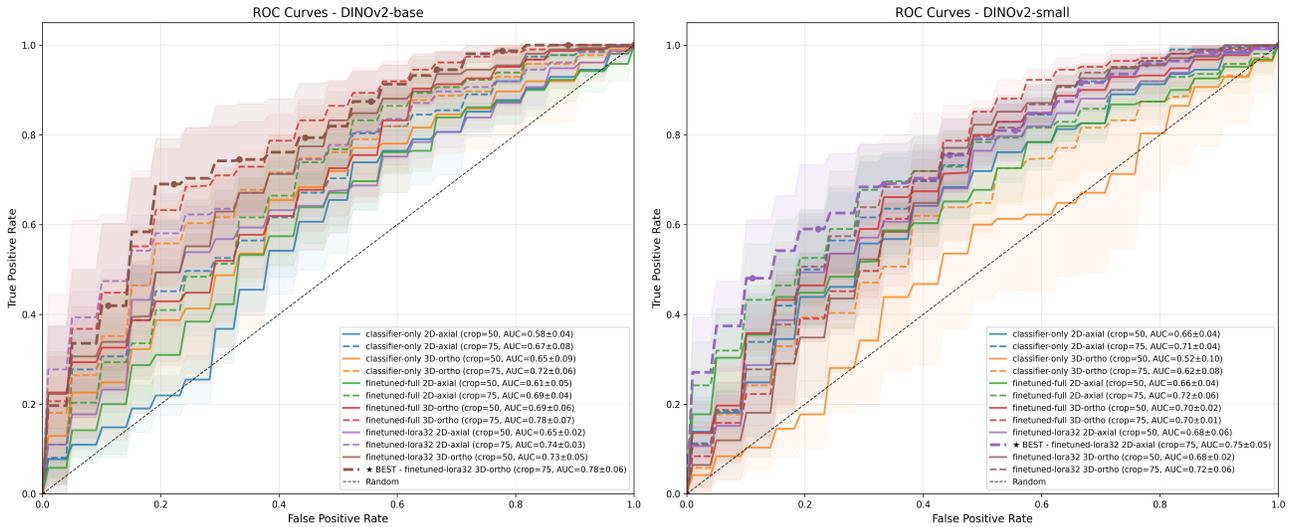

Figure 3. Receiver Operating Characteristic (ROC) curves for DINOv2-base (left) and DINOv2-small (right), comparing 2D-axial vs. 3D-orthogonal imaging and 50 vs. 75 mm³ crop sizes. Each curve corresponds to a different training strategy (NFT, FFT, or LoRA), with shaded regions indicating ±1 standard deviation across five folds. The area under the curve (AUC) is shown in parentheses.

We randomly partitioned the dataset of 221 samples into a hold-out test set (83 samples, 37.6%) and a training-validation set (138 samples, 62.4%). The 138 training-validation samples were further divided into five folds for cross-validation. In each round, 4 folds (110 samples) were used for training while the remaining fold (28 samples) was used for validation in a round-robin fashion. This approach allocated 49.8% of total data for training, 12.7% for validation, and 37.6% for testing.

All models were trained using the AdamW optimizer [13] with a class-imbalance-adjusted cross-entropy loss, a batch size of 8, and early stopping [14]. To enable a fair comparison among different fine-tuning strategies, three learning rates were employed: 1e-3 for No Fine-Tuning (NFT), 1e-4 for LoRA Fine-Tuning, and 1e-6 for Full Fine-Tuning (FFT). The best-performing checkpoint in each fold was selected based on validation performance and subsequently evaluated on the independent 83-sample hold-out test set. Final performance metrics are reported as the mean (± standard deviation) across the five cross-validation folds.

## IV. RESULTS

Tables III–VI present a comparative evaluation of five architectures—3D ResNet-10 [3], DINOv2-base, DINOv2-small, SwinV2-base, and SwinV2-small— trained on either 2D axial or 3D orthogonal inputs, with 50 mm³ or 75 mm³ crops and three fine-tuning strategies (No Fine-Tuning [NFT], Full Fine-Tuning [FFT], or Low-Rank Adaptation [LoRA]).

ROC-AUC serves as the primary performance metric in all experiments. In Table III, DINOv2-base with 3D orthogonal inputs and a 75 mm³ crop reached **0.79 ± 0.081** ROC-AUC under full fine-tuning and **0.783 ± 0.069** under LoRA, surpassing the 2D axial results ($\leq 0.728$). These configurations also outperformed the 3D ResNet-10 baseline (ROC AUC of $0.762 \pm 0.022$). Table IV compares various DINOv2-small models, where 2D axial inputs produced ROC-AUC values between 0.60 and 0.69, with LoRA and NFT approaching the upper bound of 0.69, while the 3D ResNet-10 reference achieved $0.762 \pm 0.022$. Table V indicates that SwinV2-base with 2D axial inputs and LoRA at 75 mm³ produced an ROC-AUC of about **0.735 ± 0.05**, whereas switching to 3D orthogonal inputs yielded approximately 0.70, still below the 3D ResNet-10 [3] benchmark of $0.762 \pm 0.022$. Table VI shows that SwinV2-small reached $0.747 \pm 0.03$ with 2D axial data and LoRA at 75 mm³, which was again slightly below the 3D ResNet-10 maximum of 0.762. Models trained on 3D orthogonal inputs generally yield higher ROC-AUC and F1 score, reflecting better recall and moderately high precision, whereas larger crops (75 mm³) tend to improve ROC AUC overall, although 50 mm³ remains competitive in certain settings. DINOv2 base which has a larger parameter count with 3D orthogonal data and LoRA or FFT reaches **0.78–0.79**, surpassing the 3D ResNet 10 baseline by a modest margin. LoRA often matches or closely approximates full fine tuning, whereas NFT consistently underperforms.

These findings underscore the benefit of 3D orthogonal information which provides more 3D context, particularly when combined with parameter efficient tuning approaches, for the detection of post SBRT lung injury.

In addition to these tabular comparisons, Figures 3–5 present the ROC curves for selected configurations of DINOv2 and SwinV2 models across different crop sizes and fine-tuning strategies. Each curve plots the Receiver's Operating Characteristic, displaying the model's trade-off between True Positive Rate and False Positive Rate at various decision thresholds, with the area under each curve (AUC) reported in parentheses.

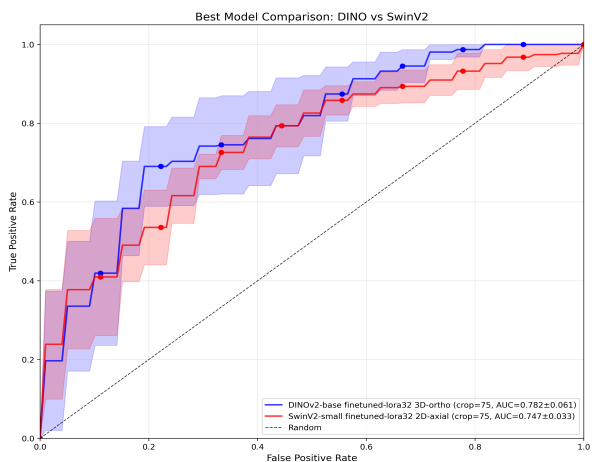

Figure 4. Comparison of best-performing DINOv2-base (3D orthogonal) and SwinV2-small (2D axial), both fine-tuned with LoRA. Shaded regions around the ROC curves represent ±1 standard deviation across 5-fold cross-validation. The diagonal line indicates random chance, and the area under each curve (AUC) is shown in parentheses.

These plots visually confirm the performance trends highlighted in Tables III–VI: 3D orthogonal inputs typically push the curves closer to the top-left corner (indicating higher True Positive rate and lower False Positive Rates for the classifier), and larger crop sizes often yield a more favorable AUC. Notably, the LoRA and Full Fine-Tuning configurations for DINOv2-base consistently achieve higher curves than the no-fine-tuning baselines, underscoring the benefits of parameter-efficient or fully adaptive training for detecting post-SBRT lung injury.

### A. Results from data for small nodules and within 3 months post SBRT

The initial three months following Stereotactic Body Radiation Therapy (SBRT) pose a significant challenge in detecting Radiation-Induced Lung Injury (RILI) due to the subtle radiographic manifestations and incomplete fibrotic tissue formation [3]. Early identification of RILI is critical for timely intervention and improved patient outcomes. Table VII evaluates the model's performance on scans obtained within this early post-treatment window, demonstrating a decline in all metrics due to the inherent difficulty of this subset. In follow-up CT scans, fibrotic regions are more pronounced in patients treated for larger nodules (>2.5 cm), making RILI detection more straightforward for clinicians [3]. To assess model performance in more challenging cases, Table VIII also analyzes nodules ≤2.5 cm, where fibrotic regions are less distinct due to smaller treatment volumes. Even within this subset, the model exhibits robust performance, achieving reliable results with a 75 mm³ input size.

## V. CONCLUSION

In this study, we demonstrated the potential of large vision models in diagnosing Radiation-Induced Lung Injury (RILI) following Stereotactic Body Radiation Therapy (SBRT) using follow-up CT scans. Our model effectively identified RILI across different post-treatment phases and nodule sizes, achieving robust performance with a ROC-AUC of **0.79±.081** and an F1 score of **0.861±.0.02** (outperforming 3D CNN model [3] which had ROC-AUC 0.762±0.22 and F1 score 0.789±0.04). Notably, it maintained predictive utility in challenging scenarios, including early post-SBRT scans and smaller nodules (≤2.5 cm).

Future work may include incorporating multi-scale ROI analysis to better capture lesion variability, leveraging Explainable AI (XAI) to improve interpretability and clinical trust, and fine-tuning on larger lung disease datasets and exploring multi-task learning in order to further enhance generalizability.